\documentclass[11pt,a4paper]{article}
\usepackage[hyperref]{emnlp2020}
\usepackage{times}
\usepackage{latexsym}
\usepackage{graphicx}
\usepackage[ruled]{algorithm2e}
\usepackage[OT1]{fontenc}

\usepackage{microtype}

\aclfinalcopy 

\title{DiDi's Machine Translation System for WMT2020}

\author{Tanfang Chen, Weiwei Wang, Wenyang Wei, \\
\textbf{Xing Shi, Xiangang Li, Jieping Ye, Kevin Knight}\\
AI Labs, DidiChuxing\\
  \texttt{\{chentanfang, wangweiweiwill, weiwenyang,}  \\ 
  \texttt{xingshi, lixiangang, yejieping, kevinknight\}@didiglobal.com}
  } 

\date{}

\begin{document}
\maketitle
\begin{abstract}
This paper describes DiDi AI Labs' submission to the WMT2020 news translation shared task. We participate in the translation direction of Chinese$\rightarrow$English. In this direction, we use the Transformer as our baseline model, and integrate several techniques for model enhancement, including data filtering, data selection, back-translation, fine-tuning, model ensembling, and re-ranking. 
As a result, our submission achieves a BLEU score of $36.6$ in Chinese$\rightarrow$English.
\end{abstract}

\section{Introduction}

We participate in the WMT2020 news translation shared tasks in Chinese $\rightarrow$ English direction. For this translation direction, we train several variants of Transformer \cite{vaswani2017attention} models on the provided parallel data enlarged with synthetic data from monolingual data.  We experiment with several techniques proposed in the past translation tasks and adopt effective ones as components of our system.

Our data preparation pipeline consists of data filtering, data augmentation, and data selection. For data filtering, we filter sentence pairs based on  language model scoring, alignment model scoring, etc. For data augmentation, we experiment with iterative back-translation \cite{sennrich2016improving,edunov2018understanding} methods and iterative knowledge distillation \cite{freitag2017ensemble} methods. We leverage source-side monolingual data by applying iterative knowledge distillation, and target-side monolingual data by back-translation methods, including greedy search, beam search, and noised beam search. 
For data selection, we select an in-domain corpus with N-grams language models and binary classifiers. A tri-gram token-level language model and a bi-gram character-level language model are introduced for English and Chinese respectively. Out-of-domain sentences which have similar scores as in-domain sentences are chosen. We also treat  data selection as a text classification problem, and use BERT \cite{devlin2019bert} as the basic classifier.
In this way, we collect a corpus of high-quality in-domain training data, which improves translation performance significantly.

To enhance a single model, we use several variants of Transformer, including Transformer with relative position attention \cite{shaw2018self}, Transformer with larger feedforward inner (FFN) size ($8,192$ or $15,000$), and Transformer with reversed source. We then ensemble these models with adequate model diversity and data diversity to further improve the performance. 

Domain conflicts influence the translation performance significantly. For example, there exist differences between written English and spoken English. Usually, a model cannot do the best in all domains due to the conflicts. In this work, we propose to obtain domain information with unsupervised clustering and exploit this information for translation. Specifically, we partition the training data, dev data, and test data into different clusters, and translate each cluster part of the test set with the model fine-tuned on the corresponding training set. Exploiting domain information helps improve the translation significantly. Details will be discussed in Section \ref{sec:system_feat}.

This paper is structured as follows: 
Section \ref{sec:model_feat} describes variants of Transformer we used in the competition. 
In Section \ref{sec:system_feat}, we introduce several techniques for model enhancement, including data filtering, back-translation, fine-tuning, model ensembling. 
Section \ref{sec:exp_res} presents experimental settings, results and analysis.
Finally, in Section \ref{sec:conclusion} we draw a brief conclusion of our work in the WMT2020.

\section{Model}
\label{sec:model_feat}

\subsection{Transformer}
The Transformer adopts a sequence-to-sequence structure, using stacked encoder and decoder layers of self-attention. Encoder layers consist of a self-attention layer followed by a feed-forward layer. Decoder layers consist of a masked self-attention layer, an encoder-decoder attention layer, and a feed-forward layer to incorporate source information and generate texts.
The residual connections \cite{he2016deep} and layer normalization \cite{ba2016layer} are introduced in the encoder and decoder layers   for better convergence.
In contrast to recurrent neural networks, the Transformer implicitly leverages relative and absolute position information in its structure. The Transformer introduces position encoding based on sinusoids in its inputs to incorporate position information.

In the competition we use Transformer Big as the baseline model, in which both the encoder and decoder have $6$ layers, the number of heads is $16$, the hidden size is $1,024$, and the feedforward inner (FFN) size is $4,096$.

\subsection{Transformer with Relative Position Attention}

The original Transformer leverages position information by taking absolute positional embeddings as inputs and does not explicitly capture the information in its structure. Thus the original Transformer cannot leverage position information efficiently. Here we used relative positional embeddings in the self-attention mechanism proposed in \citet{shaw2018self} for the encoder layers and decoder layers. We do an ablation study and find that the model with relative positional embedding has faster convergence and better performance than Transformer Big.
We adopt Transformer with relative position attention as a basic architecture in the final ensemble model.

\subsection{Transformer with Larger FFN Size}

Since increasing the model size can help improve the performance on the NMT tasks, we experiment with Transformer with a larger embedding dimension, FFN size, number of heads, and number of layers. We find that using a larger FFN size ($8,192$ or $15,000$) gives a reasonable improvement in the performance while maintaining a manageable network size. We adopt a Transformer with FFN size of $8,192$ and a Transformer with FFN size of $15,000$ as basic models in the final ensemble model, which has a larger inner dimension of feed-forward network than Transformer Big. Since Transformer with a larger FFN size is more likely to overfit, we set the dropout rate from $0.1$ to $0.3$ and use a label smoothing rate of $0.2$.

\subsection{Transformer with Reversed Source}

We reverse the source sentences of the bilingual corpus and train a Transformer with source reversed. In this way, the model can learns a different meaning of the positional embeddings, which helps capture the source sentences from a different perspective. Viewing source in a reversed order provides another kind of model diversity and data diversity and presents positive effects in the final model ensemble.

\section{System Overiew}
\label{sec:system_feat}

\subsection{Data Filtering}
\label{sec:data_filter}


Previous works \cite{sun2019baidu, xia2019microsoft, guo2019kingsoft} show that the translation performance improves as the quality of parallel corpus improves. We filter the training bilingual corpus with the following schemes:

\begin{itemize}
\item Normalize punctuation with Moses scripts
\item Filter out the sentences longer than $120$ words or sentences including a single word more than $40$ characters.
\item Filter out the sentences which contain HTML tags or duplicated translations.
\item Filter out the sentences whose languages detected by fastText\footnote{https://github.com/facebookresearch/fastText} \cite{joulin2017bag} are not identical to the translation direction.
\item Filter out the sentences whose alignment scores obtained by fast-align\footnote{https://github.com/clab/fast\_align} \cite{dyer2013simple} are low.
\item Filter out the sentences whose n-gram scores from language models are low.
\item Filter out the sentences whose length ratio between the source and target are not in range of $1:3$ and $3:1$
\end{itemize}


In this paper, we also filter out noisy sentence pairs with the translation acceptability filter proposed in \cite{zhang2020parallel}. Specifically, we feed the sentence pair $(s, t)$ into multilingual BERT, which accepts two-sentence input due to its next-sentence prediction objective. Instead of using the [CLS] token representation, we use a Convolutional Neural Network (CNN) layer that takes the BERT output and generates the final representation of the pair. Our experiments show that using CNN layer pooling achieves marginal gains over [CLS] pooling. We use the softmax probability as the degree of parallelism and filter the sentences.
The translation quality of the model boosts with the data filtering strategies.

\subsection{Large-scale Back-Translation}

The provided monolingual data contains a certain amount of noise, in which noise may affect the translation quality implicitly. Therefore, we adopt the data filtering schemes described in Section \ref{sec:data_filter}. 

Previous work \cite{edunov2018understanding} shows that leveraging the back-translation mechanism on the large-scale monolingual corpus can help improve the translation quality. \citet{edunov2018understanding} investigates several methods to generate synthetic source sentences, including greedy search, beam search, sampling top-K outputs, adding noise to beam search output, and adding noise to input sentences.

\begin{itemize}
\item Both greedy search and beam search are approximate algorithms to identify the maximum a-posteriori (MAP) output, i.e. the sentence candidate with the largest estimated probability given an input. This leads to less rich translations and is particularly problematic for text generation tasks such as back-translation.
\item Sampling top-K method selects the k most likely tokens from the output distribution, re-normalizes, and samples from this restricted set. This method is a trade-off between MAP and unrestricted sampling.
\item Adding noise to input sentences or beam search outputs can help improve the quality and robustness of the translation.
\end{itemize}

We experiment with the above methods and observe that language pairs with abundant parallel corpus like Chinese $\rightarrow$ English obtain obvious improvement with beam search and adding noise.  In our back-translation scheme, we add noise to input sentences, and use a beam search to produce the synthetic sentences. In particular, we delete words, replace words by a filler token and swap words according to a random permutation with the probability of 0.05.

\citet{zhang2018joint} proposed an iterative joint training of the source-to-target model and target-to-source model for the better quality of synthetic data. Specifically, in each iteration, the target-to-source model is responsible for generating synthetic parallel training data for the source-to-target model using the target-side monolingual data. At the same time, the source-to-target model is employed for generating synthetic bilingual training data for the target-to-source model using the source-side monolingual data. The performance of both the target-to-source and source-to-target model can be further improved iteratively. We stop the iteration when we can not achieve further improvement.

Since there are amounts of genres in both parallel and synthetic data, we adopt a language model to divide data into a coarse domain-specific corpus. We train multiple language models on different types of monolingual data (News crawl, Gigaword, etc.), and score the sentences with the language models. We select the top $600$K sentences for each domain. In the final submission, we adopt an iterative joint training scheme and train models on both bilingual and synthetic data of different genres to improve translation quality.

\subsection{Knowledge Distillation}


Alternate knowledge distillation \cite{hinton2015distilling, freitag2017ensemble} and ensemble iteratively is adopted in the competition to further boost the performance of a single model. We simply use an ensemble model as the teacher model and boost the single student model by data augmentation.
In our experiments, we use Transformer Big, Transformer with relative position, Transformer with larger FFN size, and Transformer with reversed source as basic models. For each model type, we ensemble other model types as the teacher model to boost the model performance. For example, the ensemble model of a Transformer with relative position, a Transformer with larger FFN size, and a Transformer with reversed source are adopted as a teacher model to improve the performance of a Transformer Big.



Considering that distillation from a poor-quality teacher model is likely to hurt the student network and thus results in an inferior performance, we selectively use distillation in the training process. In our experiments, we filter out data according to the sentence-level BLEU scores whose English translations lower than 28.

\subsection{In-domain Data Selection and Fine-tuning}
\label{sec:finetune}
Domain adaptation plays an important role in improving the performance towards given test data. A practical method for domain adaptation is training on the large-scale data and then fine-tuning on the in-domain data \cite{luong2015stanford}. We select the small in-domain corpus with several approaches, including N-grams language model similarity and binary classification.

\paragraph{N-grams:} We adopt the algorithm proposed in \citet{duh2013adaptation, axelrod2011domain}, which selects sentence pairs from the large out-of-domain corpus that are similar to the in-domain data. In our work, we train a tri-grams token-level language model for English and a bi-grams character-level language model for Chinese. We use the parallel texts as the out-of-domain corpus and all available test sets in the past WMT tasks and News Commentary as the in-domain corpus. We score the sentence pairs with bilingual cross-entropy differences as follows:
\begin{equation}
    \small
    CE(H_{I-SRC}, H_{O-SRC})+CE(H_{I-TGT}, H_{O-TGT})
\end{equation}
where we denote out-of-domain corpus as $O$,  in-domain corpus as $I$.  $H_{I-SRC}$ denotes language models over the source side and $H_{I-TGT}$ denotes language models over the target side on in-domain data.  $H_{O-SRC}$ denotes language models over the source side and $H_{O-TGT}$ denotes language models over the target side on out-of-domain data.  CE denotes the cross-entropy function which evaluates the differences between distributions.

Finally, we sort all sentence pairs and select the top 600K sentences with the lowest scores to fine-tuning the parameter of the model.

\paragraph{Binary Classification:} 
We also treat in-domain data selection as a text categorization problem. There are two categories: in-domain (1) and out-of-domain (0).
We use the pre-trained language model BERT as the basic classifier. For the fine-tuning data, all available newstest data and News Commentary are regarded as positive data, and randomly sampled data from the large-scale corpus are regarded as negative data. Then BERT is exploited to score the sentence pairs. We sort all sentence pairs and select the top 600K sentences with the highest scores as fine-tuning data.


All the in-domain data obtained by the above methods are adopted to fine-tuning the single model and provide about a 2 BLEU scores improvement.

\subsection{Model Ensemble}

Ensemble learning is a widely used technique in the real-world tasks, which provides performance improvement by taking advantages of multiple single models. In neural machine translation, a practical way of the model ensemble is to combine the full probability distribution over the target vocabulary of different models at each step during sequence prediction. 
We experiment with the max, avg, and log-avg strategies, and find the log-avg strategy achieves the best performance. We implement a model ensemble module in OpenNMT\footnote{https://github.com/OpenNMT/OpenNMT-tf} \cite{opennmt}. In our experiments, we observe that simply enlarging the size of ensemble models does not necessarily improve translation performance. However, brute-force search of all models is prohibitively expensive and unrealistic. As the number of models increases, the decoding of the ensemble will take more time than a single model and exceed the limits of computer resource capacity. Therefore, we adopt a greedy model ensemble algorithm \cite{li2019niutrans} as shown in Algorithm \ref{alg:ensemble}.

\begin{algorithm}[h]
\caption{An simple ensemble algorithm based on greedy search} 
\label{alg:ensemble}
\LinesNumbered
\KwIn{a model list $\Omega_{cand}$ sorted by the scores on development data.}
\KwOut{a final model list $\Phi_{final}$}
\For{all combination of 2 models that model $\in$ top-8 models}{
obtain translation by ensemble decoding  and evaluate with BLEU score\;
}
{Choose the best 2 model combination as the initial $\Phi_{final}$\;}
\While{there is tiny improvement as the model number increases}{
choose one single model from the rest of $\Omega_{cand}$ to the $\Phi_{final}$ which performs better when combined with $\Phi_{final}$\;
}
\end{algorithm}

Since model and data diversity are important factors for an ensemble system, we train diverse models with different initialization seeds, different parameters, different architectures, and different training data sets. All the models are fine-tuned to achieve superior performance.


\subsection{Domain Style Translation}
Translation performance differs in different topic domains. For intuitive explanation, we take native style and translation style as an example, and our topic domains are generated by using unsupervised clustering, not limited to these two styles. Native style and translation style are much different. A single model cannot do the best in both styles. For the Chinese $\rightarrow$ English task in WMT 2017 and 2018, the source side of both dev set and test set are composed of two parts: documents created originally in Chinese (translation style) and documents created originally in English (native style). For the Chinese$\rightarrow$English task, if the Chinese sentences are created from native Chinese corpus, then the corresponding English sentences are in translation style, so the model fine-tuned on these parallel sentences helps with translation style. Similarly, if the English sentences are created from native English corpus, the model fine-tuned on these sentences helps with native style. Previous work \cite{sun2019baidu} shows exploiting translation style and native style achieves much better performance.
In our work, we classify sentences into different topic categories (not limited to translation style and native style), and translate each specific part of the test set with the model fine-tuned on the corresponding training set.

\paragraph{Domain Label:} We use pre-trained BERT models to extract [CLS] vector as the sentence embedding and obtain two clusters by K-Means clustering. We use the cluster id as the domain label.
\paragraph{Domain Classification:} Pre-trained BERT models are fine-tuned as a text classification task, based on the source and target side with the domain label we defined above. In this way, we can select several fine-tuning data w.r.t. different topic domains.
\paragraph{Decoding Stage:} Since the test data is composed of a mixed-genre data, we first classify the domain of each sentence in the test set and obtain the probabilities corresponding to each domain. Then we apply a weighted ensemble method to integrate NMT models. Specifically, when computing the output probability of the next word, we multiply the output probability in each domain-specific translation model with the corresponding domain probability of each sentence.

\subsection{Re-ranking}

We obtain n-best hypotheses with an ensemble model and then train a re-ranker using k-best MIRA \cite{cherry2012batch} on the validation set. K-best MIRA works with a batch tuning to learn a re-ranker for the n-best hypotheses. The features we use for re-ranking are:
\begin{itemize}
\item Length Features: length ratio and length difference between the source sentences and hypotheses
\item NMT Features: scores from the ensemble model
\item Language Model Features: scores from multiple n-gram language models
\end{itemize}

\begin{table*}[thbp]
\small
\begin{tabular}{l|c|c|c|c}
\hline
                                                                                                    & Transformer Big & \begin{tabular}[c]{@{}c@{}}Transformer with \\ relative position attention\end{tabular} & \begin{tabular}[c]{@{}c@{}}Transformer with \\ larger FFN size\end{tabular} & \begin{tabular}[c]{@{}c@{}}Transformer with \\ reversed source\end{tabular} \\ \hline
baseline                                                                                            & 26.01           & 26.23                                                                                   & 26.12                                                                       & 26.08                                                                       \\
+ data augmentation & 27.02           & 27.03                                                                                   & 27.13                                                                       & 26.69                                                                       \\
+   In-domain data finetuning                                                                       & 29.33           & 29.49            & 29.62 & 29.18                                                   \\ \hline
+ model   ensemble                                                                                  & \multicolumn{4}{c}{29.72}                                                                                                                                                                                                                                             \\\hline
+ domain   style weighted                                                                           & \multicolumn{4}{c}{31.77}                                                                                                                                                                                                                                             \\\hline
+ reranking*                                                                                         & \multicolumn{4}{c}{\textbf{31.86}}                                                                                                                                                                                                                                             \\ \hline
\end{tabular}
\caption{BLEU evaluation results on the WMT 2018 Chinese $\rightarrow$ English test set (* denotes the submitted system)}
\label{tab:zh2en_wmt18}
\end{table*}

\begin{table}[thbp]
\small
\begin{tabular}{lc}
\hline
                                                        & newstest19 \\\hline
baseline                                                & 26.19                \\
+ data augmentation & 27.45                \\
+ In-domain data finetuning                             & 37.23                \\
+ model ensemble                                        & 37.64                \\
+ domain   style weighted 				& 38.59			\\
+ reranking                                             & \textbf{38.99}    \\\hline
\end{tabular}
\caption{BLEU evaluation results on the WMT 2019 Chinese $\rightarrow$ English test set}
\label{tab:zh2en_wmt19}
\end{table}

\section{Experiments and Results}
\label{sec:exp_res}

\begin{table}[thbp]
\small
\begin{tabular}{lll}
\hline
             & newstest18 & newstest19 \\ \hline
NEU \cite{li2019niutrans}          & 30.9          & 34.2          \\
MSRA \cite{xia2019microsoft}    & 30.9          & \textbf{39.3}          \\
Baidu \cite{sun2019baidu} & 31.83         & 38            \\
ours         & \textbf{31.86}         & 38.99         \\ \hline
\end{tabular}
\caption{Comparison with related work on the WMT 2018 and 2019 Chinese $\rightarrow$ English test set}
\label{tab:zh2en_sota_cmp}
\end{table}

\subsection{Experiment Setup}
Our implementation of the Transformer models is based on the version 2.3.0 of OpenNMT-tf. We use Transformer Big as a basic model. Transformer Big has $6$ layers in both encoder and decoder respectively, where each layer consists of a multi-head attention sublayer with $16$ heads and a feed-forward sublayer with inner dimension $4096$. The word embedding dimensions and the hidden state dimensions are set to $1024$ for both encoder and decoder. In the training phase, the dropout rate $P_{dropout}$ is set to $0.1$. Variants of Transformer described in Section \ref{sec:model_feat} are adopted in the competition. 

In the training phase, we use cross entropy as the loss function and apply label smoothing of $0.1$. We use Adam \cite{kingma2014adam} as our optimizer, with parameters settings $\beta_{1}=0.9$, $\beta_{2}=0.98$ and $\epsilon=10^{-8}$. The initial learning rate is set to $10^{-4}$ for training and $10^{-5}$ for fine-tuning. 
The models are trained on $4$ GPUs for about $500,000$ steps. Each model learns from data randomly sampled from the whole corpus, including bilingual data, synthetic data from back-translation, and synthetic data from knowledge distillation. Models used in iterative back-translation and knowledge distillation are trained for $200,000$ steps. We validate the model every $1,000$ steps on the development data and save the checkpoints with the best BLEU scores. After training, we average the last 10 checkpoints for every single model of the general domain. 

In the fine-tuning phase, we use the averaged model obtained in the training phase as pre-train weights for domain models, and train with in-domain data selected as in Section \ref{sec:finetune} for $10,000$ steps without early stop. After fine-tuning, we average the last 10 checkpoints for every single model of the specific domain. 

For evaluation, we adopt the cased BLEU scores calculated with SacreBLEU \cite{post-2018-call}.

\subsection{Pre-processing and Post-processing}
In pre-processing, we conduct data filtering, tokenization, subword encoding.  For Chinese sentences, we use the DiDi tokenizer for tokenization. For English data, we do punctuation normalization and use Spacy\footnote{https://github.com/explosion/spaCy} tokenizer for tokenization. We filter parallel sentences as described in Section \ref{sec:data_filter}. Finally, we collect a preprocessed bilingual training data consisting of 10M parallel sentences and 20M synthetic sentences. We adopt subword encoding for Chinese $\rightarrow$ English. Specifically, we learn a BPE with $40$K merge operations, in which $37.8$K and $27.8K$ subword tokens are adopted as Chinese and English vocabularies separately.

In the post-processing phase, we conduct unknown (UNK) words  replacement, de-tokenization, punctuation, and numerals normalization.  UNK words are simply removed in the sentences. We use the Moses scripts to true-case and de-tokenize the English translations.

\subsection{Chinese $\rightarrow$ English}
We adopt methods in Section \ref{sec:system_feat} for Chinese $\rightarrow$ English task.
Firstly we  adopt techniques of iterative back-translation and knowledge distillation for generating synthetic parallel data based on monolingual data.
We combine the synthetic data and bilingual data as the training data and randomly split training data into 6 portions and do experiments to obtain 3 most effective portions. 
We train several models  with different initialization seeds, different training datasets, and different architectures with the sampled synthetic data and bilingual data. 
In this way, we obtain models with diversity. 
After that, we fine-tune the model with different in-domain data. 
Next, we do the model ensemble by exploiting the translation domain style and choose the best model on development data as the final submission. Here we use WMT 2018 test set and WMT 2019 test set as our development data.
Finally, we adopt several re-ranking and post-processing methods to obtain the final submission.

Table \ref{tab:zh2en_wmt18} shows the results on WMT 2018 test data of Chinese $\rightarrow$ English. 
As shown in the table, data augmentation with iterative back-translation and knowledge distillation consistently improve the BLEU score. 
Fine-tuning with selected in-domain corpus plays an important role in our system, which helps achieve improvement about more than a 2 BLEU score. 
We observe that ensemble with log-avg strategy achieves slight improvement, which may be caused by the conflicts between different topic domains. To alleviate domain conflicts, we incorporate the domain style information, which achieves 2.15 improvement over the best single model.
We also observe a relatively slight improvement with re-ranking. The reason may be that we use the training data to train both the re-ranker and the NMT models, which produces similar scores while dealing with the same sentences. Similar conclusions can be drawn from Table \ref{tab:zh2en_wmt19}.

Table~\ref{tab:zh2en_sota_cmp} shows the BLEU comparisons with related works on the WMT 2018 and WMT 2019 test sets. From the table, we observe that our system achieves the best performance on the WMT 2018 test set and the second best performance on the WMT 2019 test set. This demonstrates the effectiveness of the proposed system.

In our final submission, the model is an ensemble of 6 models, including $2$ Transformer, $1$ Transformer with relative position attention, $2$ Transformer with larger FFN size, and $1$ Transformer with reversed source. We do translation with beam size=$10$ and length penalty=$1.4$. Finally, we achieve a cased BLEU score of $36.6$ in WMT 2020 Chinese $\rightarrow$ English competition.

\section{Conclusion}
\label{sec:conclusion}

In this paper, we present our NMT systems for WMT2020 news translation shared tasks in Chinese $\rightarrow$ English translation direction. Our final system achieves substantial improvement  over baseline systems by integrating the following techniques:
\begin{enumerate}
\item Data filtering
\item Data augmentation, including iterative back-translation, knowledge distillation, etc.
\item Fine-tuning with in-domain data
\item Model ensemble and leverage domain topic information
\end{enumerate}
As a result, our submitted system achieves a 36.6 BLEU score in the Chinese $\rightarrow$ English direction of WMT 2020 news translation shared tasks.


\bibliographystyle{acl_natbib}
\bibliography{emnlp2020}
\end{document}